\PassOptionsToPackage{nopatch=showhyphens}{microtype}
\documentclass[acmlarge]{acmart}

\makeatletter
\newcommand{\confshort}{\acmConference@shortname}
\newcommand{\conffull}{\acmConference@name}
\newcommand{\confdate}{\acmConference@date}
\newcommand{\confloc}{\acmConference@venue}
\AtBeginDocument{
  \fancypagestyle{firstpagestyle}{
    \fancyhead{}%
    \fancyfoot[C]{}%
  }
  \fancyhf{}
  \fancyhead[LO]{\@headfootfont\shorttitle}%
  \fancyhead[RE]{\@headfootfont\@shortauthors}%
  \fancyhead[LE]{\@headfootfont\footnotesize \confshort, \confdate, \confloc}%
  \fancyhead[RO]{\@headfootfont\footnotesize \confshort, \confdate, \confloc}%
  \fancyfoot[C]{}%
}
\makeatother

\copyrightyear{2026}
\acmYear{2026}
\setcopyright{cc}
\setcctype{by}
\acmConference[FAccT '26]{The 2026 ACM Conference on Fairness, Accountability, and Transparency}{June 25--28, 2026}{Montreal, QC, Canada}
\acmBooktitle{The 2026 ACM Conference on Fairness, Accountability, and Transparency (FAccT '26), June 25--28, 2026, Montreal, QC, Canada}
\acmDOI{}
\acmISBN{}

\usepackage{booktabs}
\usepackage{multirow}
\usepackage{tabularx}
\usepackage{array}
\usepackage{siunitx}
\usepackage{float}
\usepackage{tikz}
\usepackage[table]{xcolor}
\usetikzlibrary{arrows.meta,positioning,shapes.multipart,shapes.geometric}

\newcommand\CorpusSize{131}
\newcommand\GeocodedCount{130}
\newcommand\UniqueCountriesClean{36}
\newcommand\RegionNorthAmerica{38}
\newcommand\RegionEurope{27}
\newcommand\RegionGlobal{16}
\newcommand\RegionAfrica{16}
\newcommand\RegionMultiRegion{11}
\newcommand\RegionAsia{11}
\newcommand\RegionLatinAmerica{7}
\newcommand\RegionOceania{5}
\newcommand\RegionCombinedCount{65}
\newcommand\RegionCombinedPercent{49.6}
\newcommand\TopCountryUS{35}
\newcommand\TopCountryUK{16}
\newcommand\TopCountryCanada{13}
\newcommand\TopCountryKenya{7}

\newcommand\TopFourCountryCount{71}
\newcommand\KnownStartYears{131}
\newcommand\StartYearPresent{131}
\newcommand\StartYearMin{2006}
\newcommand\StartYearMax{2026}
\newcommand\StartAfterEighteen{124}
\newcommand\StartAfterTwentyOne{96}
\newcommand\TierCommunityLed{39}
\newcommand\TierCoDesign{26}
\newcommand\TierGovernance{22}

\newcommand\TierParticipatoryAudit{6}
\newcommand\TierPublicConsultation{6}

\newcommand\TierCoGovernance{3}
\newcommand\TopTierCount{87}
\newcommand\LifecycleProblemFormulation{85}
\newcommand\LifecycleGovernance{72}
\newcommand\LifecycleEvaluation{57}
\newcommand\LifecycleDataCollection{37}
\newcommand\LifecycleDeployment{33}
\newcommand\LifecycleDesign{24}

\newcommand\LifecycleModelTraining{16}

\newcommand\LifecycleModelDevelopment{9}
\newcommand\UniqueDomains{107}
\newcommand\SuffixOrg{46}
\newcommand\SuffixCom{16}
\newcommand\SuffixUk{10}
\newcommand\SuffixNet{8}
\newcommand\SuffixIo{7}

\newcommand\CityAnchorPresent{130}
\newcommand\LeadOrgPresent{131}
\newcommand\ProvenancePresent{131}
\newcommand\OfficialUrlPresent{131}
\newcommand\SourceTwoPresent{131}
\newcommand\SourceThreePresent{112}
\newcommand\PartnerOrgPresent{95}
\newcommand\EndYearPresent{29}

\newcommand\StatusActive{79}
\newcommand\StatusCompleted{19}
\newcommand\StatusPublishedCase{15}
\newcommand\StatusPilot{9}
\newcommand\StatusFunded{5}
\newcommand\StatusLegacy{4}
\newcommand\VerificationLive{88}
\newcommand\VerificationIndirect{25}
\newcommand\VerificationMixed{11}
\newcommand\VerificationPaper{7}
\newcommand\EvidenceA{98}
\newcommand\EvidenceB{8}
\newcommand\EvidenceC{25}
\newcommand\CoreCount{70}
\newcommand\CautiousCount{37}
\newcommand\ReviewCandidateCount{24}
\newcommand\ReviewCandidateBC{17}
\newcommand\ReviewCandidateBoundary{9}

\title[Voices in the Loop]{Voices in the Loop: Mapping Participatory AI}

\author{Rashid Mushkani}
\affiliation{
  \institution{Right to AI},
  \institution{Mila -- Québec AI Institute}
  \city{Montréal}
  \state{Québec}
  \country{Canada}
}
\email{rashidmushkani@gmail.com}
\renewcommand{\shortauthors}{Mushkani}

\begin{abstract}
Participatory approaches to artificial intelligence are increasingly documented across public, civic, and humanitarian settings, but evidence about how participation is organized remains fragmented.
This paper reports on the construction of an open repository and interactive atlas of participatory AI initiatives, using \CorpusSize{} records harmonized from Maga\~na and Shilton's Trustworthy AI corpus, and additional audited cases from research and practice.
We contribute three elements. First, we specify a reproducible protocol for discovery, vetting, harmonization, geocoding, provenance tracking, and release-based publication of participatory AI records.
Second, we report corpus-level patterns in geography, participation tiers, lifecycle loci, organizational form, verification status, and remaining documentation gaps. Documented initiatives remain concentrated in a small number of countries, while participation is most often coded at problem formulation, evaluation, and governance rather than model development or training. Third, we show how the atlas operationalizes a design and governance framework for participatory-by-default AI infrastructures through versioned releases, record-linked issue and annotation channels, schema feedback workflows, and redaction or restricted-disclosure requests. The atlas is intended to support comparative research, policy learning, and community scrutiny through a living inventory that can be updated, contested, and reused.
\end{abstract}



\begin{document}

\maketitle

\noindent\textbf{Project website:} \url{https://www.aiatlas.wiki/}

\section{Introduction}

Artificial intelligence systems are developed and deployed in contexts where decisions are contested, impacts are unevenly distributed, and knowledge about local conditions is held by actors outside development teams.
In response, public agencies, civic organizations, and researchers have adopted participatory methods intended to incorporate affected stakeholders into the design, evaluation, and governance of AI-enabled systems.
Such efforts draw on traditions in participatory design and democratic governance that treat participation as a question of decision influence rather than consultation alone \cite{arnstein1969ladder,schuler1993participatory,fung2006varieties,robertson2012participatory,Mushkani2026AIPluralismIndex}.
Within AI research and practice, participatory approaches have been discussed as a route to improve problem formulation, data practices, and accountability, while also raising concerns about extractive engagement and ``participation-washing'' \cite{sloane2022participation,birhane2022power,irani2019chasing,Mushkani2025RightToAI}.
In this paper, our focus is participation \emph{for} AI: participatory arrangements that shape the design, deployment, or oversight of an AI system. We include ``AI for participation'' only when the AI system itself is governed through such mechanisms; tools that merely support meetings, consultation, or translation without affecting an AI system's own design or governance are out of scope.

Public sector adoption of machine learning has expanded the range of settings where these concerns arise.
Algorithmic tools are used to allocate services, target inspections, and shape administrative priorities, including in urban governance and social service delivery.
Prior work documents how data-driven systems can reinforce existing inequalities when institutional constraints and political choices are treated as technical parameters \cite{eubanks2018automating,selbst2019fairness}.
As a result, governance responses combine demands for transparency about algorithmic systems with demands for participation in decisions about whether, how, and under what conditions such systems should be used \cite{wieringa2020account,kitchin2017thinking}.
Participation in this setting is not limited to user experience design; it includes institutional mechanisms such as oversight boards, labor representation, community data governance, and channels for contesting automated decisions.

Despite the growth of participatory claims, the landscape of participatory AI remains difficult to navigate.
Documentation is distributed across project websites, academic publications, government portals, and grey literature, with heterogeneous terminology and inconsistent reporting of basic metadata.
The result is a fragmented evidence base that hinders comparative research, impedes learning across sectors, and makes it difficult for communities and policymakers to identify relevant precedents.
In urban governance, the absence of systematic visibility has motivated the development of algorithm registers and transparency standards that publish structured information about algorithmic tools used by public bodies \cite{amsterdam2022playbook,helsinkiAIregister,ukATRS,AlgorithmicTransparency}.
Related concerns occur in humanitarian and civic technology, where short project cycles, shifting funders, and limited archival capacity contribute to loss of institutional memory and to limited comparability across pilots \cite{berditchevskaia2021participatory,meier2015digital}.

This paper addresses these limitations by constructing an open repository and interactive atlas of participatory AI initiatives assembled from public documentation available through early 2026.
The repository is designed as a living infrastructure rather than a static catalog.
It is intended to support ongoing additions, corrections, and disputes about how participatory claims are operationalized in practice.

Our empirical focus is \CorpusSize{} projects. The dataset harmonizes records from the 95-project Trustworthy AI corpus reported by Magaña and Shilton \cite{magana2025frameworks}, along with additional audited searches across research and practice documentation. Because records can enter through more than one discovery path, we treat the corpus as an integrated collection rather than partitioning it by source.

Across this corpus, we curate geospatial, documentary, and participation-specific metadata to support a global map view and comparative analysis of where participation is located in the AI lifecycle, how strong the underlying evidence is, and which records remain contested or provisional. We also examine what can be learned from the corpus.

Participation tier, participants, methods, and lifecycle stages are coded for every record, shifting the main analytical bottleneck from the absence of participation fields to the uneven evidentiary quality and archival durability of public documentation. Using this dataset, we identify four empirical patterns: concentration in a small set of jurisdictions; the dominance of community-led, co-design, and governance-oriented forms of participation; a long tail of provenance domains; and persistent gaps in partner constellations, closure timelines, and third-source corroboration.

The contribution of the paper is threefold.
We specify a reproducible protocol for discovery, vetting, and harmonization of project records, including deduplication, geocoding, provenance tracking, and release-based archiving.
We then report empirical patterns in the corpus, emphasizing geographic concentration, participation tiers, lifecycle loci, organizational forms, verification status, and remaining documentation gaps.
Finally, we propose a design and governance framework for participatory-by-default AI infrastructures.
The framework synthesizes lessons from public algorithm registers, collective crisis intelligence pilots, indigenous data governance, and community-maintained open science \cite{carroll2020care,kukutai2016indigenous,berditchevskaia2021participatory,bowker2000sorting}.
Rather than treating participation as an optional layer added to technical work, the framework treats participation as a default property of infrastructures that mediate data collection, model development, and public accountability. The contribution is therefore both infrastructural and analytical: we provide a reusable, versioned corpus and atlas, but we also use that corpus to make claims about the current conditions under which participatory AI can be compared, contested, and critically evaluated.

The remainder of the paper is organized as follows.
Section~\ref{sec:related} reviews prior work relevant to participatory AI mapping and documentation.
Section~\ref{sec:definition} defines the operational scope used for mapping.
Section~\ref{sec:protocol} presents the mapping protocol and repository design.
Section~\ref{sec:results} reports corpus-level patterns and documentation gaps.
Section~\ref{sec:governance} proposes governance and design requirements for participatory-by-default infrastructures.
Section~\ref{sec:discussion} discusses implications and limitations, followed by conclusions.

\section{Background and related work}
\label{sec:related}

Mapping participatory AI draws on theory from participatory design and democratic governance, and on practical ways to document and compare sociotechnical systems.
We review four areas: (1) participation in design and governance, (2) participatory approaches in AI/ML, (3) documentation and evaluation artifacts, and (4) registers, inventories, and systematic mapping in public technology governance.

Participatory design emerged from labor and workplace democratization movements that treated technology design as a locus of power, negotiation, and situated expertise \cite{schuler1993participatory,simonsen2012participatory}.
Its methods emphasize in-context inquiry, iterative prototyping with stakeholders, and explicit attention to whose interests are represented.
In governance research, participation is often assessed by influence, representation, and institutional form.
Arnstein's ladder distinguishes tokenistic consultation from delegated power \cite{arnstein1969ladder}, while Fung's typology differentiates participation by who participates, how they deliberate, and what authority they hold \cite{fung2006varieties}.
Public administration work highlights how these mechanisms operate under legal, budgetary, and capacity constraints, including when participation is mandated without commensurate resources \cite{nabatchi2012public}.

Within AI and machine learning, participatory methods have been argued to matter across the lifecycle, including problem formulation, data practices, evaluation, and monitoring \cite{birhane2022power,sloane2022participation}.
A central critique is that ``participation'' can become extractive labor or validation theater when communities supply data or legitimacy without meaningful decision influence or reciprocal benefit \cite{sloane2022participation,irani2019chasing}.
Human-centered AI research also emphasizes tensions between local specificity and the pressures of scale, especially for systems deployed across jurisdictions with different norms and institutional capacities \cite{selbst2019fairness}.
In parallel, public interest technology and design justice emphasize participation as a strategy to redistribute design authority and align system goals with community-defined priorities \cite{costanza2020design,dIgnazio2020data}.
Together, these debates motivate mapping that distinguishes participation claims from documented mechanisms, and that treats missing documentation as an empirical finding rather than a nuisance.
Two recent lines of work provide particularly direct context for our contribution. Delgado et al. synthesize the theoretical foundations of the ``participatory turn'' in AI design and show that current practice varies substantially in how much agency it grants stakeholders \cite{delgado2023participatory}. We draw directly on that distinction between participation claims and decision influence when defining inclusion criteria. Kawakami et al.'s Situate AI Guidebook provides a co-designed toolkit for early-stage, multi-stakeholder deliberation around proposed public sector AI systems \cite{kawakami2024situate}. That work is prospective and intervention-oriented; by contrast, our atlas documents already public initiatives as an empirical case base that can be compared, contested, and revisited over time.

Documentation artifacts enable comparison across diverse projects.
Proposals such as datasheets and model cards standardize reporting about datasets/models, intended use, and performance \cite{gebru2018datasheets,mitchell2019model}; data statements formalize provenance and collection conditions in NLP \cite{bender2018data}; and algorithmic impact assessments and auditing frameworks aim to structure disclosure about risks, mitigations, and oversight \cite{reisman2018algorithmic,raji2020closing}.
These approaches inform repository schema design but typically foreground technical artifacts rather than participatory process metadata.
Evaluation traditions add relevant distinctions: value sensitive design treats values as explicit design constraints that can be elicited and negotiated \cite{friedman2019vsd}, while participatory evaluation differentiates participation-as-data-collection from shared control over evaluation questions, interpretation, and use \cite{cousins1998framing}.
For participatory AI mapping, this suggests documenting not only who was engaged, but how engagement shaped decisions and how outcomes were assessed.

Work on infrastructures and classification frames an atlas or repository as a governance instrument rather than a neutral index.
What becomes visible and actionable depends on standards, installed bases, and classification choices \cite{star1996ecology,star1999ethnography,bowker2000sorting}.
Registries enact categories and defaults that shape accountability; a ``living'' participatory AI atlas must therefore anticipate how schemas and review processes can reproduce exclusions, and must support contestation and revision over time.

Public-sector technology governance offers concrete precedents for living inventories.
Municipal algorithm registers (Helsinki and Amsterdam) publish structured descriptions of deployed systems and provide feedback channels \cite{helsinkiAIregister,amsterdam2022playbook}.
The UK Algorithmic Transparency Recording Standard provides a disclosure template covering purpose, data, oversight, and impact considerations \cite{ukATRS}, while transnational efforts aim to harmonize transparency categories across jurisdictions \cite{AlgorithmicTransparency}.
The New York City ADS Task Force report illustrates both the demand for visibility and the institutional difficulty of standardizing disclosure \cite{nycads2019report}.
While these instruments focus on systems inside public administrations, participatory AI initiatives often span civic organizations, academia, and humanitarian networks; nonetheless, register practices around provenance, versioning, and public contestation are instructive for participatory AI documentation.

A closely related line of work is Maga\~na and Shilton's systematic review of 95 global projects connecting participatory AI to trustworthy AI \cite{magana2025frameworks}. Their contribution is an analytic synthesis of frameworks, methods, and shared tasks across documented cases. Our work instead provides a reusable documentation infrastructure: a release-based atlas with harmonized project records, provenance and verification fields, participation-tier and lifecycle coding, and governance mechanisms for contestation, redaction, and schema evolution. More broadly, repositories such as Participedia demonstrate the value of large-scale, crowdsourced case libraries for democratic innovations \cite{fung2011participedia,landry2022participedia}. Our atlas draws inspiration from this infrastructural model, but differs in being AI-specific and in encoding fields that broader participation repositories typically do not, including technical modality, the locus of participation within the AI lifecycle, evidence of decision influence, provenance links, and sensitivity controls such as redaction or restricted disclosure.

Finally, systematic mapping has established protocols in evidence synthesis, specifying search strategies, inclusion criteria, and extraction templates \cite{kitchenham2007guidelines}.
Participatory AI mapping adds complications: conceptual ambiguity, uneven participation across lifecycle stages, and documentation bias or intentional non-disclosure for sensitive contexts.
We therefore treat the repository itself as an evolving artifact, whose categories and governance must remain revisable as new cases, critiques, and community feedback emerge.
Taken together, these literatures motivate the design of a documentation infrastructure that enables systematic comparison, critical scrutiny, and ongoing revision of participatory AI claims.

\section{Operational definition and scope}
\label{sec:definition}

``Participatory AI'' is used inconsistently across research and practice, so mapping requires an operational definition that is both extractable and flexible across institutional contexts.
We define a \emph{participatory AI project} as an initiative that meets two conditions: (1) it develops, deploys, or maintains an AI-enabled system or workflow, where AI refers to machine learning or related statistical inference used to automate, support, or structure decisions; and (2) it includes a \emph{documented} mechanism through which affected stakeholders, community representatives, or frontline practitioners can influence decisions about problem definition, data governance, model development, deployment conditions, or oversight.

This definition excludes projects that only use public data or conduct user testing without decision influence, and it distinguishes concrete participation mechanisms from aspirational ethical claims.
Consistent with participation research, the key criterion is the existence of decision points where stakeholder input can shape outcomes \cite{arnstein1969ladder,fung2006varieties}.
Participation is treated as a feature of governance arrangements rather than of any single workshop, interface, or engagement activity.
This boundary also distinguishes participation \emph{for} AI from AI \emph{for} participation. We focus on the former. We include the latter only when public documentation shows that the AI system itself was shaped or governed through participatory mechanisms.

Because documentation practices vary by sector, participation may be visible through different traces.
Public algorithm registers can provide feedback channels and structured templates \cite{helsinkiAIregister,ukATRS}.
Citizen science platforms often operationalize participation via volunteer data collection and distributed validation \cite{bonney2009citizen}.
Indigenous language technologies may center participation through community authority over data licensing and use conditions \cite{carroll2020care,jones2025kaitiaki}.
The repository therefore stores participation as a harmonized curatorial field linked to provenance, verification status, and review state, so that inclusions can be audited rather than taken as self-evident.

We bound the mapping in three ways.
\textbf{Temporal scope:} the current analysis uses the early 2026 release snapshot, with known project start years ranging from \StartYearMin{} to \StartYearMax{} rather than a fixed intake window.
\textbf{Geographic scope:} global coverage, with location recorded at the granularity supported by public sources.
\textbf{Substantive scope:} emphasis on public, civic, and humanitarian settings.
Projects whose primary contribution is an AI tool for facilitating deliberation, translation, or consultation are only included when the tool's own design or governance was itself participatory.

To enable comparison across heterogeneous cases, extraction records five dimensions: participation locus, stakeholder roles, decision influence, documentation artifacts, and institutional setting.
Appendix Table~\ref{tab:participation-dimensions} lists the public indicators used for each dimension.
These dimensions align the repository with AI lifecycle and governance research while remaining grounded in evidence that can be verified from public documentation.

Participation mechanisms also cluster into recurring, non-exclusive families that differ in how influence is exercised and what traces they leave.
Table~\ref{tab:participation-families} provides the families used as a conceptual guide for extraction and later coding.

\begin{table}[t]
\centering
\caption{Families of participation mechanisms observed in documented participatory AI initiatives and their typical evidence traces.}
\label{tab:participation-families}
\begin{tabularx}{\linewidth}{@{}p{0.36\linewidth}p{0.56\linewidth}@{}}
\toprule
Participation mechanism family & Typical evidence traces in public documentation \\
\midrule
Co-design and co-production & Workshop reports, partnership descriptions with community organizations, iterative prototyping, or field trials \\
Distributed contribution & Volunteer recruitment and data collection protocols, community annotation practices, validation interfaces, contributor governance pages \\
Oversight and contestation & Register entries, complaint and appeal channels, oversight board descriptions, public responses to feedback \\
Data stewardship and licensing & Community data governance statements, licensing terms, consent and redaction policies, data access committees \\
\bottomrule
\end{tabularx}
\end{table}

This operationalization structures the mapping protocol in Section~\ref{sec:protocol} and motivates the governance framework in Section~\ref{sec:governance}, which treats documentation and contestability as core features of participatory-by-default infrastructures.

\section{Mapping protocol and repository design}
\label{sec:protocol}

This section describes the data sources, the mapping protocol used to construct project records, and the repository and atlas design decisions that support ongoing maintenance.

\subsection{Data sources}

The current release integrates two input streams.
The first is the 95-project Trustworthy AI corpus reported by Magaña and Shilton \cite{magana2025frameworks}.
The second is an audited set of additional cases drawn from research and practice, selected to broaden coverage of governance-oriented participation mechanisms, including public algorithm registers, volunteer coordination infrastructures, and community-governed language technologies.

\subsection{Discovery and vetting protocol}

The protocol aims to be reproducible, while acknowledging that search results and web content change over time.
Discovery uses a combination of keyword searches, network expansion, and targeted domain queries.
Keyword searches identify projects that self-describe using terms such as ``co-design'', ``citizen science'', ``community governance'', ``participatory'', and ``public register'' in proximity to AI-related terms.
Network expansion follows links from known civic technology organizations, humanitarian innovation labs, and municipal digital service units.
Targeted queries focus on domains where participation mechanisms have established precedents, including public sector transparency registers and community-run data stewardship initiatives.
Because the public-interest dataset and audited searches were conducted primarily through English- and German-language web sources, the protocol is better understood as a structured recovery procedure than as a language-neutral discovery method. This matters for interpretation: over-representation of European and North American cases may reflect both underlying activity and differential recoverability through those documentation ecologies.

Vetting applies the operational definition in Section~\ref{sec:definition}.
A project must have an identifiable AI-enabled component and a documented participation mechanism.
Documentation must be publicly accessible at the time of extraction and must provide enough information to identify at least a country-level location and a responsible organization or consortium.
Projects are excluded when participation claims are purely aspirational, when documentation is limited to press releases without operational details, or when the only evidence of participation is passive data collection without stakeholder influence.
Borderline cases are retained when participation can be inferred from governance artifacts, for example when a project record is part of a municipal register with documented oversight or when community data licensing terms are explicitly specified.

\subsection{Extraction template}

For each included project, the repository stores a canonical project name, a primary provenance URL that identifies the source used during extraction, and a set of harmonized metadata fields.
Descriptive attributes include location, lead organization, organization type, activity status, application domain, AI modality, and start and end years where available.
The dataset also codes participation-specific fields, including participation tier, participants, participation methods, and AI lifecycle stages.
Evidence fields include up to three corroborating source URLs, free-text participatory evidence notes, verification status, evidence grade, and review status.
Extraction therefore distinguishes descriptive codings from the documentary basis on which those codings rest.

Harmonization resolves differences in naming and formatting across sources.
Name normalization resolves superficial differences such as whitespace and punctuation.
Country names are normalized to reduce duplication caused by variant spellings.
Where the same initiative appears in multiple sources, the repository retains one canonical record and preserves alternate links for audit.
The repository also retains completeness information for fields that remain uneven across public documentation, notably partner organizations, end years, and multi-source corroboration.

\subsection{Geocoding, location resolution, and data sensitivity}

Geocoding is performed using available location information in project documentation.
When a city and country are available, the record is geocoded to a representative point coordinate for that locality.
When only a country is available, the record is geocoded to a country-level reference coordinate and flagged accordingly.
In the current release, \GeocodedCount{} of \CorpusSize{} records have mappable coordinates; one global record is intentionally left ungeocoded.
Coordinates are treated as approximate anchors for atlas visualization, not as precise operational sites.
This approach avoids implying that participation occurs uniformly within a city or country, and it limits the risk of exposing sensitive operational locations in humanitarian and conservation contexts.

The repository does not store personal data about individual participants.
Stakeholder roles are recorded at the level of organizations or collectives when they are publicly documented.
For projects involving Indigenous communities or other groups with collective governance requirements, the repository prioritizes documentation that specifies community authority and consent conditions \cite{carroll2020care}.
The governance framework in Section~\ref{sec:governance} further proposes mechanisms for redaction and restricted disclosure requests.

\subsection{Provenance}

Provenance tracking is implemented as a mandatory field.
Each record includes at least one URL pointing to the source used for extraction, and many records include additional corroborating links.
The repository is designed for versioned publication with archival snapshots so that analyses and citations can be tied to stable releases.
This design is aligned with documentation practices in algorithm registers, where transparency relies on traceable records that can be contested and revised \cite{ukATRS,amsterdam2022playbook}.
In addition to provenance, the repository stores a change log for edits, including the reason for change and the contributor identity when available.
Releases are intended to be periodic rather than continuously mutating public state, so citations can point to stable snapshots even as corrections accumulate between releases.
All numeric values in Section~\ref{sec:results} refer to a frozen release snapshot rather than to the continuously updated live atlas. Alongside each release, we therefore archive the CSV snapshot, schema version, manifest, and derived JSON and GeoJSON artifacts used for the paper's counts, so that readers can reproduce the reported tables and figures against the same state of the corpus.

\subsection{Atlas interface}

\begin{figure*}[t]
\centering
\includegraphics[width=0.98\textwidth]{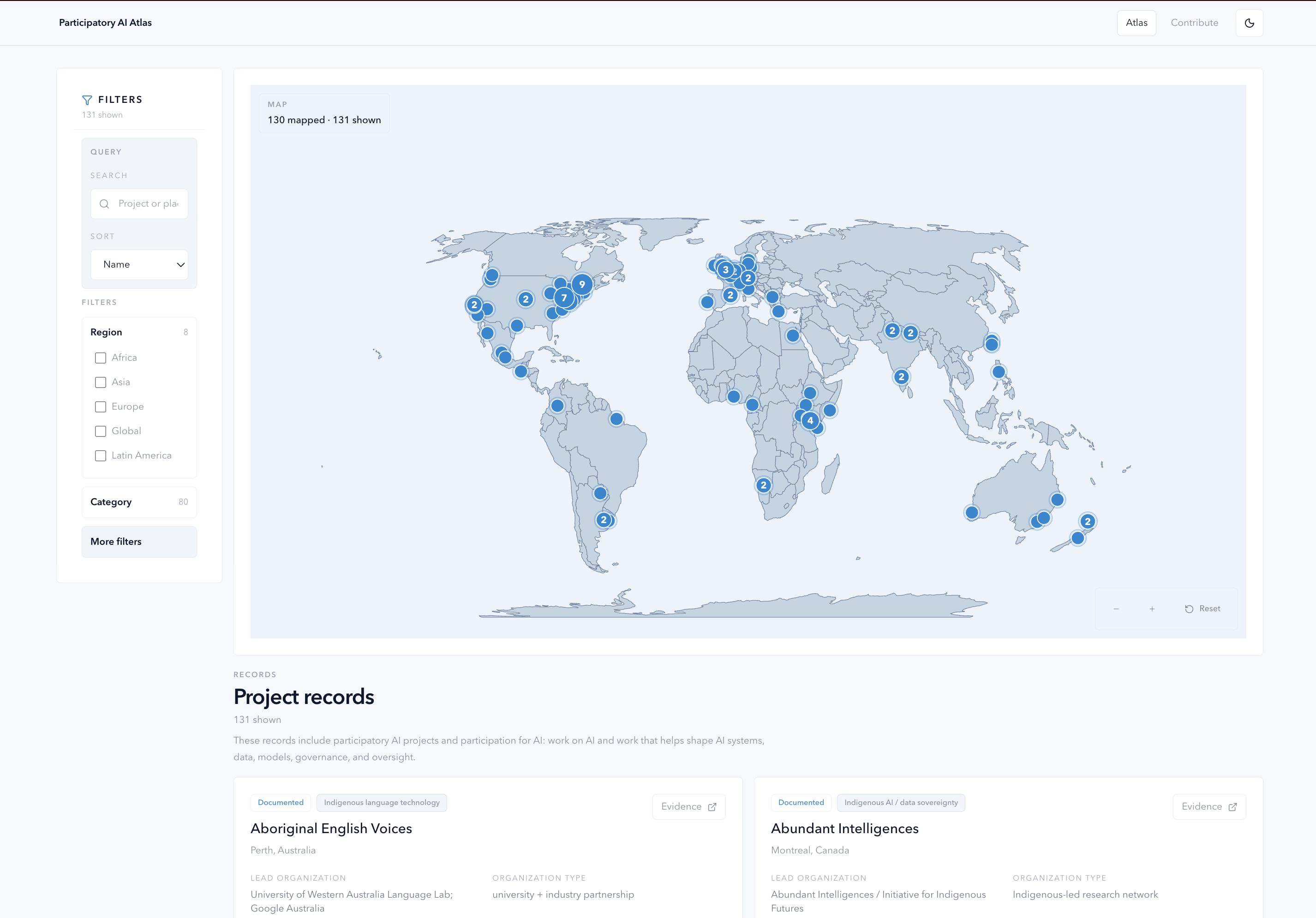}
\caption{Atlas interface showing the explorer view, map markers, and filter controls for the \CorpusSize{}-project release.}
\Description{A screenshot of the atlas explorer. A left sidebar contains search and region filters. The main panel shows a world map with project markers. A header at the top includes navigation items for the atlas and contribution workflow.}
\label{fig:atlas-ui}
\end{figure*}

The atlas interface renders the repository as an interactive map with filtering by geography, application domain, organization type, participation tier, and review-related fields.
It supports drill-down to project pages that display provenance, evidence signals, and links to documentation artifacts.
A core design decision is to treat each record as a pointer to external documentation rather than as an authoritative representation of project practice.
The atlas therefore emphasizes transparency about what is known, what is missing, and what is contested.
In the current implementation accompanying this paper, each public record exposes provenance, documentation completeness, release version, evidence grade, and review status; stores edit history; and links to archived release artifacts. Record pages include separate channels for (i) correction or dispute submissions, (ii) contextual annotations, and (iii) redaction or restricted-disclosure requests. New records enter through a moderated intake form rather than immediate publication, while schema-change proposals are handled in a separate governance queue. These design choices operationalize participation through multiple contribution channels, contestability through record-linked issue handling, and stewardship through archived releases and disclosure controls.
Figure~\ref{fig:atlas-ui} makes these mechanisms concrete.

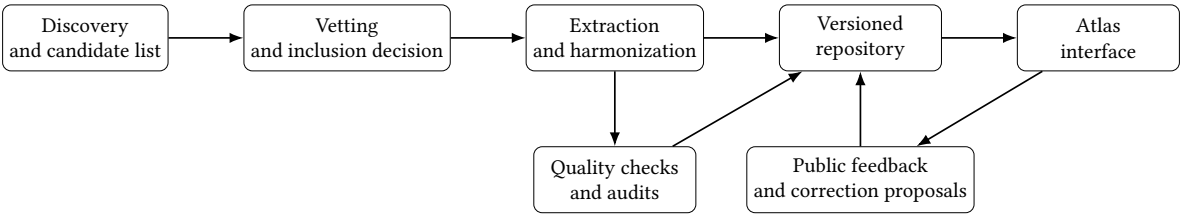
\begin{figure*}[t]
\centering
\resizebox{0.98\textwidth}{!}{%
\begin{tikzpicture}[
  node distance=1.25cm,
  box/.style={draw, rounded corners, align=center, minimum width=2.7cm, minimum height=1.1cm},
  arrow/.style={-Latex, thick}
]
\node[box] (discover) {Discovery\\and candidate list};
\node[box, right=of discover] (vet) {Vetting\\and inclusion decision};
\node[box, right=of vet] (extract) {Extraction\\and harmonization};
\node[box, right=of extract] (repo) {Versioned\\repository};
\node[box, right=of repo] (atlas) {Atlas\\interface};

\draw[arrow] (discover) -- (vet);
\draw[arrow] (vet) -- (extract);
\draw[arrow] (extract) -- (repo);
\draw[arrow] (repo) -- (atlas);

\node[box, below=of repo] (feedback) {Public feedback\\and correction proposals};
\draw[arrow] (atlas) -- (feedback);
\draw[arrow] (feedback) -- (repo);

\node[box, below=of extract] (qa) {Quality checks\\and audits};
\draw[arrow] (qa) -- (repo);
\draw[arrow] (extract) -- (qa);

\end{tikzpicture}%
}
\caption{Workflow for constructing and maintaining a living repository and atlas of participatory AI projects. The feedback loop is treated as part of the infrastructure, not as a post-publication add-on.}
\Description{A flow diagram with boxes labeled Discovery, Vetting, Extraction, Repository, and Atlas connected left to right, with additional boxes for Quality checks and Public feedback feeding into the versioned repository.}
\label{fig:pipeline}
\end{figure*}

Figure~\ref{fig:pipeline} summarizes the end-to-end workflow from discovery to atlas publication and feedback.
The workflow is designed so that external contributors can propose additions and corrections, while maintaining a review process that records disagreements and preserves provenance.

The protocol and schema are intended to be extensible.
Sections~\ref{sec:results} and \ref{sec:governance} use the corpus to identify documentation gaps and to motivate governance mechanisms that support long-term maintenance, including stewardship structures, contribution review, and policies for sensitive information.

\section{Empirical findings from the mapped corpus}
\label{sec:results}

This section reports empirical patterns in the mapped corpus. Because participation tier, participants, methods, and AI lifecycle stages are coded across all \CorpusSize{} records, the central questions concern where these codings are concentrated, the strength of the evidence supporting them, and which documentary gaps remain. When we refer below to low, medium, and high missingness, we mean fields absent in less than 30\%, between 30--70\%, and more than 70\% of records, respectively. Because the corpus combines existing datasets with audited expansion and interpretive harmonization, the descriptive results reported here should be understood as properties of the dataset rather than unbiased prevalence estimates of the full participatory AI landscape.

\subsection{Geographic distribution}

\begin{figure*}[t]
\centering
\includegraphics[width=0.98\textwidth]{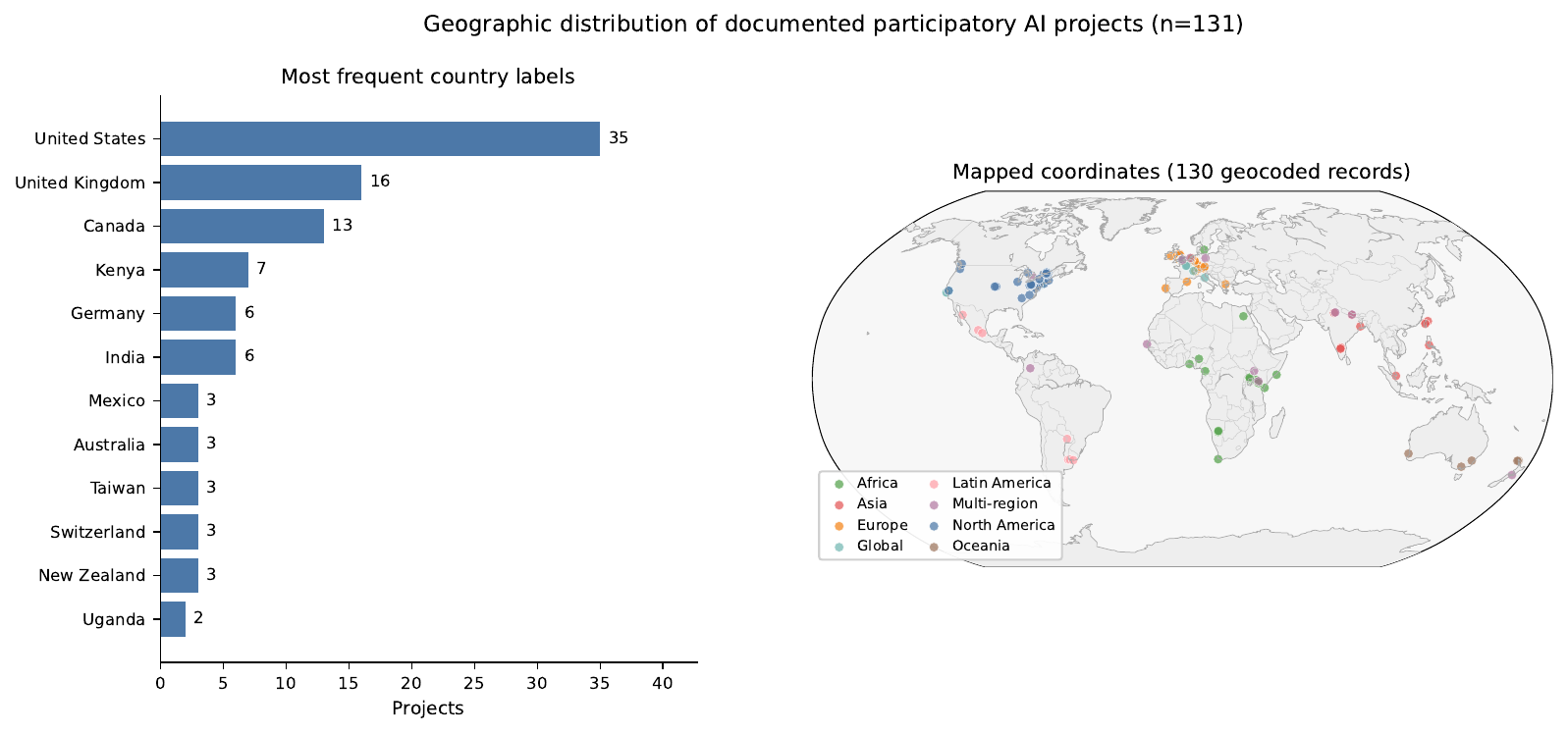}
\caption{Geographic distribution of participatory AI records (\CorpusSize{} projects). The bar chart uses normalized country labels; the map shows approximate anchors for the \GeocodedCount{} geocoded records.}
\Description{A combined figure with a horizontal bar chart of the most frequent normalized country labels and a world map with colored project markers by region. The largest bars are the United States, the United Kingdom, Canada, and Kenya. Markers cluster most densely in Europe and North America, with additional records across Africa, Asia, Latin America, and Oceania.}
\label{fig:geo}
\end{figure*}

Figure~\ref{fig:geo} shows the geographic distribution of projects using a combined view of top countries and mapped coordinates.
The corpus spans \UniqueCountriesClean{} normalized country labels, alongside global and multi-country records.
North America and Europe remain the largest regions, with \RegionNorthAmerica{} and \RegionEurope{} projects, followed by Africa (\RegionAfrica{}), Asia (\RegionAsia{}), Latin America (\RegionLatinAmerica{}), and Oceania (\RegionOceania{}).
Multi-region and global records add \RegionMultiRegion{} and \RegionGlobal{} entries.
The four largest single-country labels are the United States (\TopCountryUS{}), the United Kingdom (\TopCountryUK{}), Canada (\TopCountryCanada{}), and Kenya (\TopCountryKenya{}), which together account for \TopFourCountryCount{} of \CorpusSize{} projects.

North America and Europe together account for \RegionCombinedCount{} of \CorpusSize{} records (\RegionCombinedPercent{}\%).
The concentration should therefore not be read as a clean proxy for where participatory AI is most active in absolute terms. It captures an interaction between underlying activity, institutional visibility, and recoverability through the protocol. Public algorithm registers and other web-visible disclosure infrastructures are especially legible in Europe and North America, while cases documented only in other languages or in less indexable formats are less likely to be recovered.
An aggregate regional view of the same pattern is provided in Appendix~\ref{app:supplement} (Figure~\ref{fig:continents}).
The concentration implies that generalizations about participatory AI based on documented projects will over-represent institutional contexts and governance regimes prevalent in those regions.

\subsection{Participation tiers, lifecycle loci, and substantive domains}

Three tiers dominate the release: community-led initiatives (\TierCommunityLed{} projects), co-design (\TierCoDesign{}), and participatory governance (\TierGovernance{}).
Together they account for \TopTierCount{} of \CorpusSize{} records.
Public consultation (\TierPublicConsultation{}), participatory audit (\TierParticipatoryAudit{}), and co-governance (\TierCoGovernance{}) appear, but much less frequently.
This distribution matters because it shows that the mapped landscape is not only about interface co-design; governance-oriented and community-led arrangements constitute a substantial share of documented cases.

The same pattern appears across lifecycle loci.
Counting each stage whenever a project lists more than one, participation is most often documented in problem formulation (\LifecycleProblemFormulation{} projects), governance (\LifecycleGovernance{}), evaluation (\LifecycleEvaluation{}), data collection (\LifecycleDataCollection{}), deployment (\LifecycleDeployment{}), and design (\LifecycleDesign{}).
By contrast, model training (\LifecycleModelTraining{}) and model development (\LifecycleModelDevelopment{}) are relatively rare.
In the current release, public documentation therefore more often locates participation around agenda setting, evaluation, and oversight than around core model-building stages.
Appendix~\ref{app:supplement} provides the full participation-tier and lifecycle distributions (Figures~\ref{fig:tiers} and \ref{fig:lifecycle}).

Substantively, the most frequent exact application-domain labels concern healthcare, AI governance and trustworthy AI, AI policy and public administration, assistive technology and accessibility, and human rights documentation and accountability.
At the category level, public dialogue and governance, inclusive stakeholder engagement, public-health mental-health support, and trustworthy AI co-design are especially common.
Together, these patterns suggest that participatory AI is documented most densely in settings where legitimacy, accountability, and situated expertise are central to adoption.

\subsection{Organizational forms, activity status, and temporal profile}

Organizational form is heterogeneous, but the release is anchored in public-interest and collaborative settings.
The most common lead-organization labels are university/research lab and nonprofit/NGO (\num{10} each), standalone university and research consortium (\num{9} each), university plus civic methodology partner (\num{8}), and nonprofit/research collaboration (\num{7}).
This mix reinforces that many participatory AI initiatives operate through hybrid institutional arrangements rather than clearly bounded firms or agencies.

Status fields are available per project.
\StatusActive{} projects are marked active, \StatusCompleted{} completed, \StatusPublishedCase{} published case, \StatusPilot{} pilot, \StatusFunded{} funded, and \StatusLegacy{} legacy.
Start years are available for \KnownStartYears{} projects and range from \StartYearMin{} to \StartYearMax{}, but the corpus is heavily recent: \StartAfterEighteen{} dated records begin in 2018 or later and \StartAfterTwentyOne{} in 2021 or later.
The atlas therefore captures not only mature precedents such as registers and long-running language initiatives, but also a wave of recent governance, deliberation, and auditing efforts.
These dates should not be read as a complete time series of the field, because inclusion still depends on recoverable public documentation rather than exhaustive discovery.

\subsection{Documentation completeness, verification, and remaining gaps}

Table~\ref{tab:completeness} summarizes completeness of selected fields.
Participation mode, participants, participation methods, AI lifecycle stages, lead organization, primary provenance URL, and official project URL are coded for all \CorpusSize{} records.
The documented subset includes second corroborating sources for all records. Remaining gaps are concentrated elsewhere: third sources are present for \SourceThreePresent{} records, partner organizations for \PartnerOrgPresent{}, and end years for only \EndYearPresent{}.
By the thresholds above, third corroborating sources and partner organizations exhibit low missingness, while end years exhibit high missingness.
These gaps matter because they limit inference about institutional durability, coalition structure, and project closure.

\begin{table}[t]
\centering
\caption{Completeness of selected repository fields (\CorpusSize{} projects).}
\label{tab:completeness}
\footnotesize
\begin{tabularx}{\linewidth}{@{}p{0.62\linewidth}r r@{}}
\toprule
Field & Records & Percent \\
\midrule
City or locality anchor & \CityAnchorPresent{} & 99.2\% \\
Lead organization & \LeadOrgPresent{} & 100.0\% \\
Primary provenance URL & \ProvenancePresent{} & 100.0\% \\
Official project URL & \OfficialUrlPresent{} & 100.0\% \\
Second corroborating source & \SourceTwoPresent{} & 100.0\% \\
Third corroborating source & \SourceThreePresent{} & 85.5\% \\
Partner organizations & \PartnerOrgPresent{} & 72.5\% \\
Start year & \StartYearPresent{} & 100.0\% \\
End year & \EndYearPresent{} & 22.1\% \\
\bottomrule
\end{tabularx}
\end{table}

Completeness alone does not settle reliability.
The release differentiates verification status, evidence grade, and review state.
\VerificationLive{} records are live\_verified, \VerificationIndirect{} indirect\_verified, \VerificationMixed{} mixed\_verified, and \VerificationPaper{} paper\_verified.
Evidence grades are A for \EvidenceA{} records, B for \EvidenceB{}, and C for \EvidenceC{}.
The atlas also makes curatorial uncertainty explicit: \CoreCount{} records are marked core, \CautiousCount{} cautious, and \ReviewCandidateCount{} review\_candidate.
Review-candidate entries disproportionately carry B/C evidence grades (\ReviewCandidateBC{} of \ReviewCandidateCount{}) and are often coded as participatory governance or public consultation (\ReviewCandidateBoundary{} of \ReviewCandidateCount{}), indicating that scope boundary cases cluster in policy-facing and oversight-oriented material rather than in the most concrete co-design cases.

\begin{figure}[t]
\centering
\includegraphics[width=0.78\linewidth]{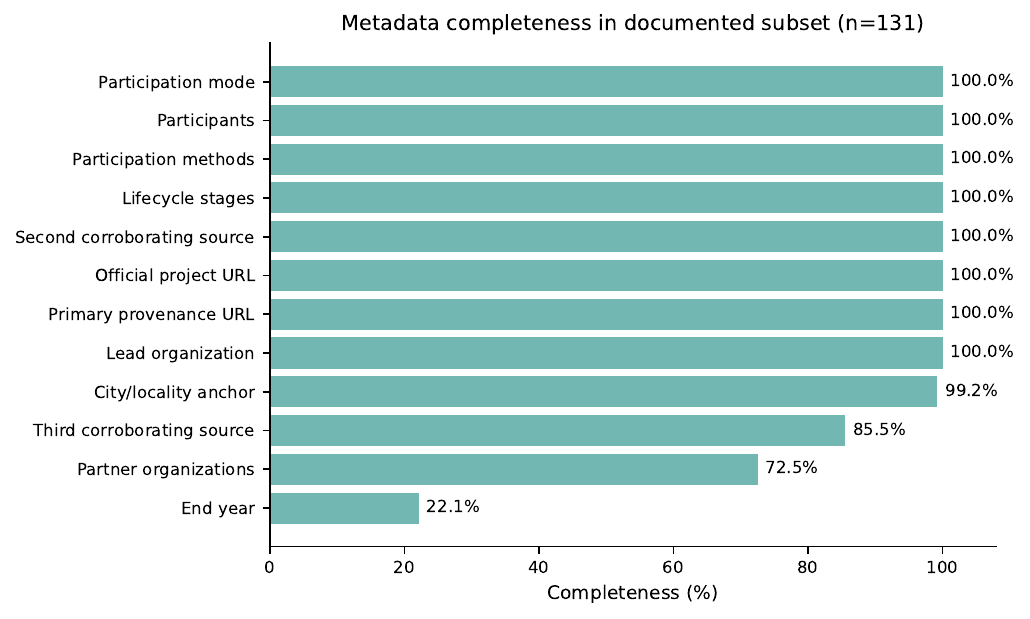}
\caption{Completeness of selected atlas fields}
\Description{A horizontal bar chart showing completeness percentages for selected fields. Lead organization, provenance URL, official URL, and participation-related fields are complete, while partner organizations and end year have substantially lower coverage.}
\label{fig:completeness}
\end{figure}

Figure~\ref{fig:completeness} visualizes completeness for selected fields.
Completeness reflects the presence of a non-empty value and does not assess the specificity or verifiability of the value.

\subsection{Provenance sources}

The corpus draws on documentation hosted across \UniqueDomains{} distinct provenance domains, indicating a long tail of organizational websites, research pages, and repositories rather than a small set of durable registries.
The most common domain suffixes in provenance URLs are \texttt{.org} (\SuffixOrg{} records), \texttt{.com} (\SuffixCom{}), \texttt{.uk} (\SuffixUk{}), \texttt{.net} (\SuffixNet{}), and \texttt{.io} (\SuffixIo{}).
This distribution reflects both organizational heterogeneity and the continuing dependence on web pages that may not provide long-term archival stability.
The long tail of \UniqueDomains{} source domains is itself a result: participatory AI is presently documented through a dispersed and often unstable web ecology rather than through a few durable registries or repositories. This helps explain why a versioned atlas matters. Without release snapshots and archived provenance, the empirical basis for comparison can disappear as organizational pages change or vanish.
Appendix~\ref{app:supplement} provides a ranked list of the most frequent provenance domains (Table~\ref{tab:provenance-domains}).

\subsection{Implications of curation and bias}

The mapped corpus should not be interpreted as a census of participatory AI. It is a curated release built from recoverable public documentation and harmonized through interpretive coding.
The atlas materially improves participation legibility by coding tier, participants, methods, and lifecycle stage for every record, but it does so through curatorial judgment that remains open to correction.
Records with lower evidence grades or review-candidate status are therefore not noise; they mark where the boundary between participation for AI and adjacent governance, deliberation, or advocacy infrastructures is hardest to draw.
Documentation bias remains visible in multiple ways.
Language, searchability, funding ecosystems, and disclosure mandates shape recoverability.
Public-sector registers and English-language research sites remain easier to surface than informal, non-indexed, or local-language initiatives.
These biases motivate the governance requirements in Section~\ref{sec:governance}, where the atlas is treated as an infrastructure that must expose not only cases but also uncertainty, disagreement, and the conditions under which records were included.
The atlas is therefore best read as a map of documented legibility. Cases with disclosure mandates, funder reporting obligations, or dedicated communications staff become easier to recover than equally consequential local initiatives documented through informal, non-indexed, or non-English channels.

\section{Design and governance framework}
\label{sec:governance}

Section~\ref{sec:results} shows that documentation of participatory AI is uneven.
A living atlas can support accountability only if participation, contestability, and stewardship are treated as core infrastructure properties.
The framework below is therefore a governance contribution.
It treats documentation, contestability, and disclosure control as infrastructural conditions for evaluating participatory AI claims.

\subsection{From project catalogs to infrastructure stewardship}

A participatory AI repository has a reflexive risk: it can reproduce extractive dynamics if communities are merely documented rather than positioned as contributors to how records are interpreted, corrected, and governed.
Public algorithm registers offer a partial model through structured disclosure and feedback channels \cite{helsinkiAIregister,amsterdam2022playbook,ukATRS}, but participatory AI initiatives often span public agencies, community organizations, and volunteer networks.
The atlas in Figure~\ref{fig:atlas-ui} is designed around this premise.
Participation is operationalized through contribution channels; contestability through record-linked corrections, disputes, and annotations; and stewardship through release governance, provenance logging, and redaction pathways.
The dataset also makes evidentiary uncertainty visible through verification status, evidence grade, and review status fields, so that inclusion does not imply equal evidentiary strength.

\subsection{Design requirements for a participatory-by-default atlas}

We propose five requirements: verifiable provenance; minimal viable participation documentation; community rights and licensing; contestability and change tracking; and interoperability with technical documentation artifacts.

\textbf{Verifiable provenance.} Each entry should include at least one stable source reference, and the atlas should preserve prior versions for audit and citation.

\textbf{Minimal viable participation documentation.} A minimally interpretable record should specify (i) lifecycle locus, (ii) participants as collectives or organizations, (iii) decision points influenced, and (iv) the mechanism translating input into decisions.
These criteria establish a minimum documentation threshold rather than a certification of participatory quality. Records that invoke participation but do not specify locus, mechanism, or decision influence are treated as documentation-insufficient, rather than being implicitly accepted as participatory.

\textbf{Community rights and licensing.} Records should represent data governance and consent conditions when relevant, and communities should be able to request redaction or restricted disclosure for sensitive information \cite{carroll2020care,kukutai2016indigenous,jones2025kaitiaki}.

\textbf{Contestability and change tracking.} Because participatory practice is described differently by project teams and stakeholders, the atlas should support issue tracking, structured disagreement, and links to third-party assessments \cite{raji2020closing,reisman2018algorithmic}.
Annotations should remain distinct from direct record edits so that critical commentary and counter-documentation do not disappear into a revised canonical entry.

\textbf{Interoperability with technical artifacts.} Where available, project records should link to datasheets, model cards, and data statements \cite{gebru2018datasheets,mitchell2019model,bender2018data}, while preserving participation metadata that those artifacts usually omit.

Table~\ref{tab:governance-elements} summarizes governance elements that operationalize these requirements.

\begin{table*}[t]
\centering
\caption{Governance elements for a participatory-by-default atlas, operationalized in the current implementation and motivated by prior infrastructures.}
\label{tab:governance-elements}
\footnotesize
\setlength{\tabcolsep}{4pt}
\begin{tabularx}{\textwidth}{@{}p{0.21\textwidth}p{0.56\textwidth}p{0.19\textwidth}@{}}
\toprule
Governance element & Operational mechanism in the atlas & Motivating precedents \\
\midrule
Contribution rights and review & Moderated intake; required provenance and minimal participation fields; review decisions logged & Open source workflows; municipal registers \cite{ukATRS,amsterdam2022playbook} \\
Contestability & Record-linked corrections and disputes; annotations separate from edits; optional third-party assessments & Comment processes; auditing \cite{helsinkiAIregister,raji2020closing} \\
Sensitive information policy & Field-level visibility plus redaction or restricted-disclosure requests for safety-sensitive contexts & Humanitarian and do-no-harm practice \cite{berditchevskaia2021participatory,meier2015digital} \\
Release governance and archiving & Periodic releases, change logs, archived manifests, and citation-ready snapshots & Register maintenance; archival practice \cite{ukATRS,star1999ethnography,bowker2000sorting} \\
Schema evolution & Separate schema-feedback queue; stewardship review of field changes; decisions recorded in release notes & Community-maintained repositories \\
Stewardship and representation & Multi-stakeholder stewardship group with regional curation and conflict-of-interest policy & Data stewardship and Indigenous governance principles \cite{carroll2020care,kukutai2016indigenous} \\
\bottomrule
\end{tabularx}
\end{table*}

\subsection{Participatory evaluation}

An atlas supports evaluation only if it links participation mechanisms to decision outcomes.
Participatory evaluation research treats evaluation criteria as governance because they define success and encode values \cite{cousins1998framing}.
The atlas can therefore help build shared scrutiny standards without becoming a certification scheme: it does not score participation quality on its own, but it can establish what must minimally be documented if a participatory claim is to be examinable. Community annotations and linked third-party audits then provide a way to record disagreement when project self-description overstates stakeholder influence.

\subsection{Implementing ``living'' participation in the atlas}

A living atlas requires durable responsibility for maintenance and channels for contribution beyond project owners.
In practice, new records and record edits enter a moderated intake queue with required provenance and minimal participation fields. Accepted changes are published through periodic releases. Each release is archived with a manifest and downloadable artifacts so that analyses remain citable.

Record-linked corrections and disputes are handled through issue threads tied to the relevant entry. Annotations are distinct from edits: they can attach contextual notes, missing documentation, or critical commentary without overwriting the source record. Schema evolution is handled separately from record maintenance through a stewardship review process documented in release notes. These mechanisms make the atlas ``living'' in two senses: the corpus can be corrected and expanded, and the schema itself can change in response to critique.

\subsection{Reducing documentation and coverage bias}

Because the atlas inherits the biases of searchable documentation, maintenance must include incentives and partnerships, not only better search queries. Multilingual submission templates and translation support can lower entry barriers; regional curators or partner networks can audit for missing cases; compensated community review can reduce extractive reporting burdens; and interoperability with public registers or existing repositories can reduce duplicate documentation work. These steps do not eliminate bias, but they make recoverability an object of stewardship rather than an invisible artifact of search ranking and language privilege.

\subsection{Case-based grounding}

Appendix Table~\ref{tab:cases} summarizes illustrative cases showing why different participation mechanisms imply different documentation needs. Across municipal registers, humanitarian pilots, Indigenous language technologies, and community-facing platforms, the common requirement is a documented link between participation mechanisms and decision influence.

\section{Discussion and limitations}
\label{sec:discussion}

The mapping and governance framework have implications for research, policy, and participatory practice.

For research, the repository enables comparative analysis of participation mechanisms across domains, jurisdictions, and organizational forms, and supports sampling and systematic reviews that do not depend on a small set of canonical cases.
It also enables work linking participation to evaluation, for example by comparing whether stakeholder-defined success criteria are documented alongside technical artifacts such as datasheets and model cards \cite{gebru2018datasheets,mitchell2019model}.
Because tier and lifecycle codings are now available across the full release, the main empirical constraint is not the total absence of participation fields, but the labor required to triangulate heterogeneous documentation and to keep those codings auditable over time.

For policy and governance, the atlas can complement algorithm registers by covering initiatives outside public administrations, including civic and humanitarian projects that still shape public decision making.
The framework suggests that transparency mandates should include participation reporting requirements, not only templates for purpose, data, and oversight \cite{ukATRS}.
This helps move discussion of participation-washing from ``was participation claimed?'' to ``what was documented about locus, mechanism, participants, and decision influence?'' That shift does not settle normative adequacy, but it creates a baseline for scrutiny.

For practice, a living atlas can help communities and developers find related efforts, reuse documentation templates, and reduce duplicated work.
Linking records to datasheets and model cards can support technical transparency, but these artifacts rarely capture decision influence or governance arrangements \cite{bender2018data}.
The practical implication is that the atlas should accumulate not only cases, but also the conditions for scrutiny: release history, disagreement, third-party commentary, and documentation of what remains unknown.

The work has limitations.
The corpus reflects projects with publicly accessible documentation, under-representing informal, safety-sensitive, and local-language initiatives.
Location fields are approximate anchors rather than full operational geographies. Furthermore, many values are harmonized curatorial summaries rather than verbatim source statements, and they remain sensitive to boundary judgments.
End years are available for only \EndYearPresent{} projects and partner organizations for \PartnerOrgPresent{}, limiting inference about project closure and coalition structure.
Start years are available across the documented subset, but start-year values still reflect what public sources make comparable rather than a field-wide timeline.
The atlas has also not yet been validated through formal user studies or through structured consultation with researchers and practitioners in underrepresented regions to determine which cases remain invisible because they are documented differently rather than not documented at all.

Addressing these limits will require consultation with regional research and practice networks, multilingual intake and review workflows, compensated community curation, crosswalks to existing public registers and repositories, and future interface evaluation with the atlas's intended users. These are governance tasks, not ancillary improvements.

\section{Conclusion}
\label{sec:conclusion}

Participatory approaches to AI are increasingly documented in public, civic, and humanitarian settings, yet their distribution and implementation remain difficult to assess from scattered public materials.
This paper reported the construction of a versioned repository and interactive atlas of \CorpusSize{} participatory AI projects.
We specified a reproducible mapping protocol, reported empirical patterns in geography, participation tiers, lifecycle loci, organizational forms, and evidence status, and identified the remaining documentary gaps that constrain comparison.
Across the release, documented initiatives remain geographically concentrated, while participation is most often located in problem formulation, evaluation, and governance rather than model development or training.
The central problem, in other words, is not only locating participatory AI initiatives, but documenting them in ways that keep claims about participation open to scrutiny, correction, and reuse.

To address these gaps, we proposed a design and governance framework for participatory-by-default AI infrastructures.
The framework treats provenance, contestability, community rights, and interoperability with technical documentation as core infrastructure requirements.
A living atlas can support research, policy learning, and collaboration only if it is itself governed as a participatory artifact, with mechanisms for contribution, annotation, and dispute resolution.
The repository and atlas provide an initial basis for such work and can be extended through community-led stewardship and alignment with emerging transparency standards.

\subsection*{Generative AI usage}
Generative AI tools were used to assist with language editing and grammar. No generative AI system was used to create the dataset or generate the empirical counts reported in the paper.

\bibliographystyle{ACM-Reference-Format}
\bibliography{0_01_references}

\clearpage

\appendix
\section{Supplementary figures and tables}
\label{app:supplement}

This appendix collects supplementary operationalization details, distributions, and ranked domain summaries that are useful for interpretation.

\subsection{Operationalization reference table}

\begin{table}[H]
\centering
\caption{Participation-related dimensions used in the repository schema. The mapping emphasizes evidence that can be verified from public documentation.}
\label{tab:participation-dimensions}
\footnotesize
\setlength{\tabcolsep}{3pt}
\begin{tabularx}{\textwidth}{@{}p{0.19\textwidth}p{0.31\textwidth}p{0.25\textwidth}p{0.17\textwidth}@{}}
\toprule
Dimension & Operationalization & Examples of evidence & Typical verification challenge \\
\midrule
Participation locus & Lifecycle stage(s) where stakeholders can influence decisions (problem framing, data governance, model development, deployment, oversight) & Project documentation describing co-design workshops, community data licensing, oversight boards, or feedback channels & Multi-stage influence is often distributed across several sources \\
Stakeholder roles & Named roles beyond developers and funders, including affected groups, service workers, community organizations, and public institutions & Governance pages, community partner lists, memoranda, worker councils & Consortium pages often blur affected groups, implementers, and funders \\
Decision influence & Indication that stakeholder input changes decisions, not only that consultation occurred & Revision history, minutes, published responses to feedback, changes to data access conditions & Influence is frequently described narratively rather than through auditable change logs \\
Documentation artifacts & Publicly available outputs that support accountability and reuse (model cards, datasheets, impact assessments, registry entries) & Links to reports, code repositories, datasets, registry records & Artifacts are dispersed across reports, repositories, and institutional pages \\
Institutional setting & Organization type(s) and administrative context (public agency, non-profit, academic, private, hybrid) & Legal base, organizational owner, funding description & Hybrid partnerships resist single-label classification \\
\bottomrule
\end{tabularx}
\end{table}

\subsection{Regional, participation, and lifecycle distributions}

\begin{figure}[t]
\centering
\includegraphics[width=0.95\linewidth]{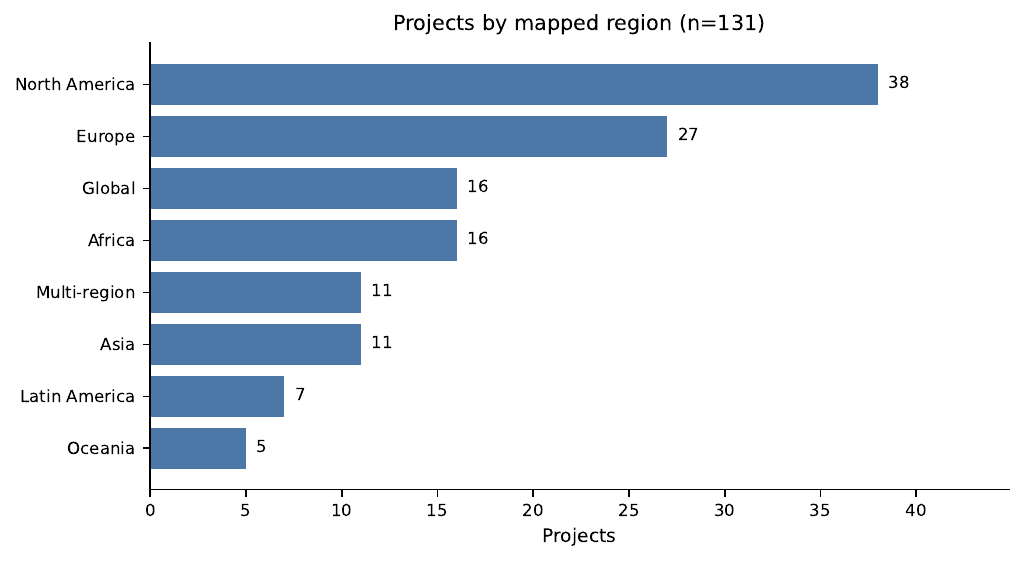}
\caption{Projects by mapped region in the March 2026 release.}
\Description{A horizontal bar chart showing counts by region. North America and Europe are the largest regional groups, followed by global, Africa, multi-region, Asia, Latin America, and Oceania.}
\label{fig:continents}
\end{figure}

\begin{figure}[t]
\centering
\includegraphics[width=0.95\linewidth]{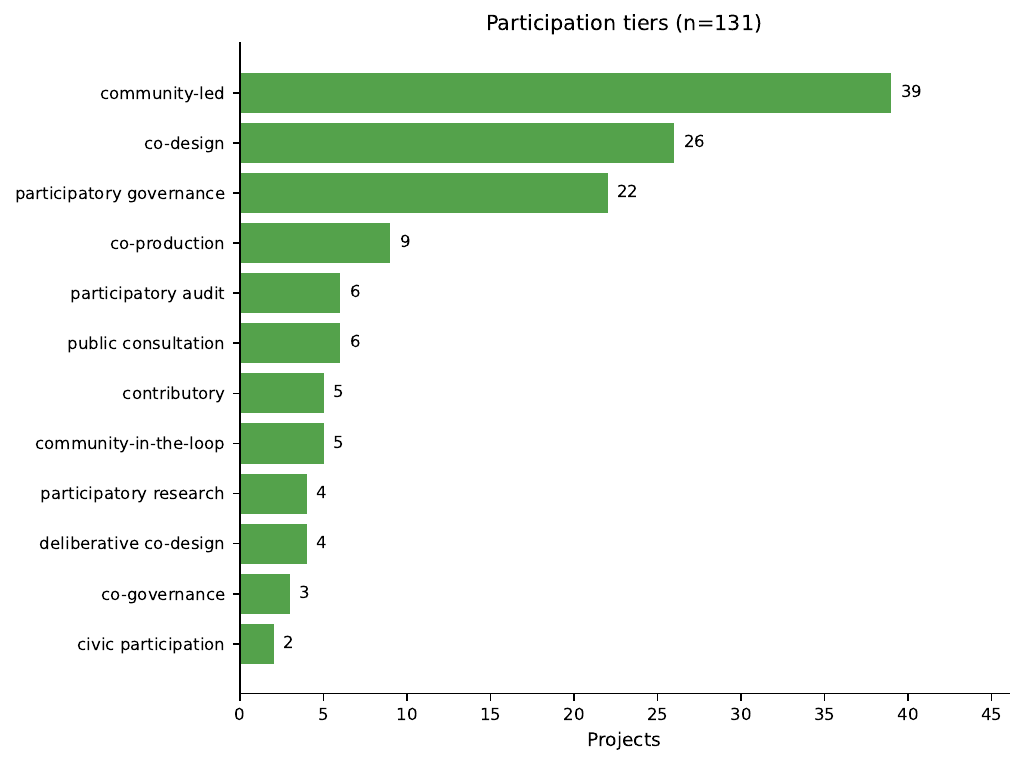}
\caption{Distribution of participation tiers across the March 2026 release.}
\Description{A horizontal bar chart showing counts by participation tier. Community-led, co-design, and participatory governance are the largest categories.}
\label{fig:tiers}
\end{figure}

\begin{figure}[t]
\centering
\includegraphics[width=0.95\linewidth]{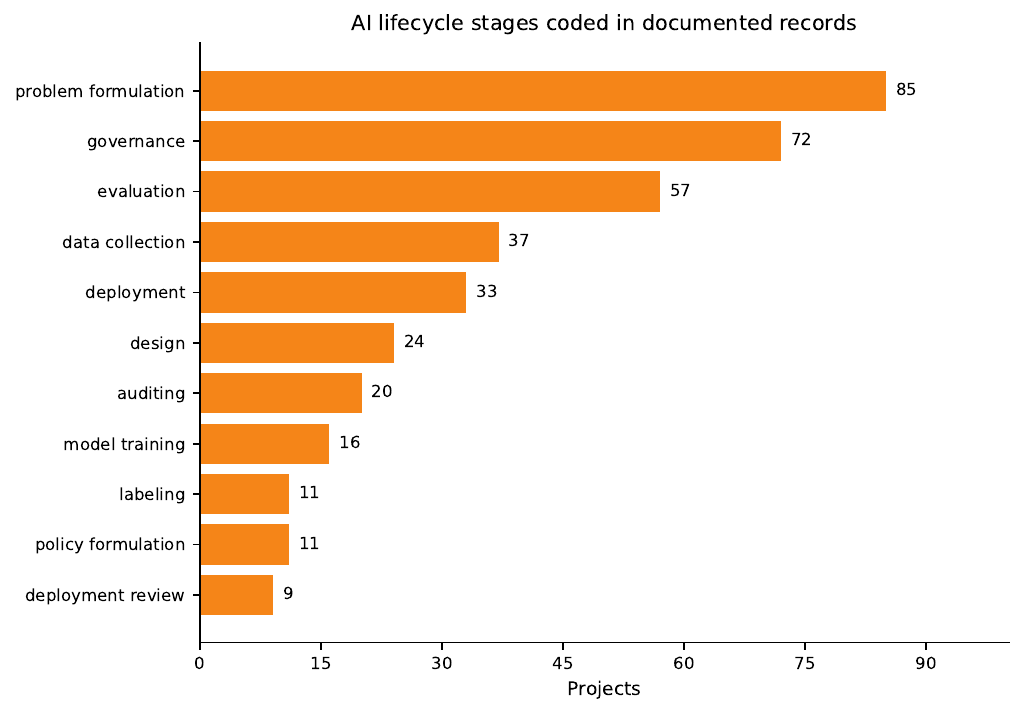}
\caption{Most frequently coded AI lifecycle stages, counting projects that list more than one stage.}
\Description{A horizontal bar chart showing the most common lifecycle stages. Problem formulation is the largest category, followed by governance, evaluation, data collection, deployment, and design, while model training and model development are much smaller.}
\label{fig:lifecycle}
\end{figure}

\subsection{Provenance domains}

\begin{table}[H]
\centering
\caption{Most frequent provenance domains in the corpus (top ten).}
\label{tab:provenance-domains}
\footnotesize
\begin{tabular}{@{}l r@{}}
\toprule
Domain & Projects \\
\midrule
\texttt{aplusalliance.org} & 5 \\
\texttt{dl.acm.org} & 5 \\
\texttt{masakhane.io} & 5 \\
\texttt{mila.quebec} & 4 \\
\texttt{moda.gov.tw} & 3 \\
\texttt{nesta.org.uk} & 3 \\
\texttt{arxiv.org} & 2 \\
\texttt{cdacnetwork.org} & 2 \\
\texttt{cndp.us} & 2 \\
\texttt{openforgood.info} & 2 \\
\bottomrule
\end{tabular}
\end{table}

\section{Illustrative cases for grounding governance requirements}
\label{app:cases}

\begin{table}[t]
\centering
\caption{Illustrative cases used to ground governance requirements for a living participatory AI atlas.}
\label{tab:cases}
\footnotesize
\setlength{\tabcolsep}{3pt}
\begin{tabularx}{\linewidth}{@{}p{0.27\linewidth}p{0.31\linewidth}p{0.34\linewidth}@{}}
\toprule
Case type & Participation mechanism documented & Atlas design and governance implication \\
\midrule
Municipal algorithm registers \cite{helsinkiAIregister,amsterdam2022playbook} & Public disclosure templates and feedback channels tied to administrative responsibility & Template-driven disclosure can raise baseline comparability; records require stable versioning and an escalation path for disputes \\
Collective crisis intelligence pilots \cite{berditchevskaia2021participatory} & Co-design with humanitarian practitioners and affected communities under operational constraints & Records require sensitivity controls, clear provenance, and guidance for what should not be publicly disclosed \\
Indigenous language technologies \cite{jones2025kaitiaki,leoni2024tehiku} & Community control of data governance and licensing conditions for model training and reuse & Records must represent authority to control and consent conditions, and must enable community redaction requests \\
Open conservation platforms & Distributed sensing and community validation linked to conservation governance arrangements & Records must distinguish operational sites from organizational bases and should support longitudinal tracking and archiving \\
Agricultural diagnosis tools for smallholders & Hybrid human mediation and feedback loops between users, agronomists, and tool developers & Records should capture how feedback alters model updates and how local expertise is represented in evaluation \\
School-home communication translation platforms & Human-in-the-loop translation workflows with user feedback and institutional adoption in schools & Records should capture deployment contexts and governance arrangements between platform providers and public institutions \\
\bottomrule
\end{tabularx}
\end{table}

\end{document}